\documentclass[conference]{IEEEtran}
\IEEEoverridecommandlockouts

\usepackage{smartdiagram}
\usepackage{cite}
\usepackage{amsmath,amssymb,amsfonts}
\usepackage{caption}
\usepackage{subcaption}
\usepackage{textcomp}
\usepackage{xcolor}
\usepackage{booktabs}
\usepackage[binary-units]{siunitx}
\usepackage{soul}
\usepackage{xcolor}
\usepackage{comment}
\usepackage[printonlyused,nohyperlinks, nolist]{acronym}
\usepackage[shortcuts,acronym,automake=false]{glossaries}
\usepackage{multirow}

\usepackage{enumitem}

\usepackage{amsmath}
\usepackage{amssymb}

\usepackage{stfloats}

\usepackage{url}

\usepackage{multirow}

\usepackage[hidelinks]{hyperref}

\usepackage[printonlyused,nohyperlinks,nolist]{acronym}
\usepackage{algorithm}
\usepackage{algpseudocode}
\usepackage[capitalise,noabbrev]{cleveref}

\usepackage[numbers,sort&compress]{natbib}

\usepackage{dsfont}

\usepackage{siunitx}
\usepackage{import}
\usepackage{color}

\DeclareUnicodeCharacter{2212}{-}

\def\BibTeX{{\rm B\kern-.05em{\sc i\kern-.025em b}\kern-.08em
    T\kern-.1667em\lower.7ex\hbox{E}\kern-.125emX}}
\begin{document}

\newacronym{v2x}{V2X}{vehicle-to-everything}
\newacronym{v2v}{V2V}{vehicle-to-vehicle}
\newacronym{v2i}{V2I}{vehicle-to-infrastructure}
\newacronym{v2p}{V2P}{vehicle-to-pedestrian}
\newacronym{v2n}{V2N}{vehicle-to-network}
\newacronym{ml}{ML}{machine learning}
\newacronym{lte}{LTE}{long-term evolution}
\newacronym{dsrc}{DSRC}{dedicated short-range communication}
\newacronym{cits}{C-ITS}{cooperative intelligent transport systems}
\newacronym{qos}{QoS}{quality-of-service}
\newacronym{tod}{ToD}{tele-operated driving}
\newacronym{mno}{MNO}{mobile network operator}
\newacronym{e2e}{E2E}{end-to-end}
\newacronym{gps}{GPS}{global positioning system}
\newacronym{rsrp}{RSRP}{reference signal received power}
\newacronym{rssi}{RSSI}{received signal strength indicator}
\newacronym{snr}{SNR}{signal-to-noise ratio}
\newacronym{mae}{MAE}{mean absolute error}
\newacronym{mse}{MSE}{mean squared error}
\newacronym{rsrq}{RSRQ}{reference signal received quality}
\newacronym{ue}{UE}{user equipment}
\newacronym{3gpp}{3GPP}{3rd Generation Partnership Project}
\newacronym{mcs}{MCS}{modulation and coding scheme}
\newacronym{cam}{CAM}{cooperative awareness messages}
\newacronym{cpm}{CPM}{collective perception messages}
\newacronym{sdr}{SDR}{software-defined radio}
\newacronym{per}{PER}{packet error ratio}
\newacronym{harq}{HARQ}{hybrid automatic repeat request}
\newacronym{rrc}{RRC}{radio resource control}
\newacronym{rrm}{RRM}{radio resource management}
\newacronym{phy}{PHY}{physical layer}
\newacronym{ul}{UL}{uplink}
\newacronym{dl}{DL}{downlink}
\newacronym{minipc}{DME}{dedicated measurement equipment}
\newacronym{tx}{TX}{transmission}
\newacronym{rx}{RX}{reception}
\newacronym[firstplural=radio access technologies (RATs)]{rat}{RAT}{radio access technology}

\title{
Berlin V2X: A Machine Learning Dataset from Multiple Vehicles
and Radio Access Technologies
\thanks{This work was supported by the Federal Ministry of Education and Research (BMBF) of the Federal Republic of Germany as part of the AI4Mobile project (16KIS1170K). The authors alone are responsible for the content of the paper. The described dataset is publicly available at \url{https://dx.doi.org/10.21227/8cj7-q373} \cite{hernangomez2022berlinv2x}.}
}

\author{\IEEEauthorblockN{Rodrigo Hernang\'{o}mez\IEEEauthorrefmark{1},
Philipp Geuer\IEEEauthorrefmark{4}, Alexandros Palaios\IEEEauthorrefmark{4}, Daniel Schäufele\IEEEauthorrefmark{1}, Cara Watermann\IEEEauthorrefmark{4},
Khawla Taleb-\\Bouhemadi\IEEEauthorrefmark{1}, Mohammad Parvini\IEEEauthorrefmark{2}, Anton Krause\IEEEauthorrefmark{2}, Sanket Partani\IEEEauthorrefmark{3}, Christian Vielhaus\IEEEauthorrefmark{5},
Martin Kasparick\IEEEauthorrefmark{1},
Daniel \\ F. K\"{u}lzer\IEEEauthorrefmark{7}, Friedrich Burmeister\IEEEauthorrefmark{2},
Frank H. P. Fitzek\IEEEauthorrefmark{5}, Hans D. Schotten\IEEEauthorrefmark{3}, Gerhard Fettweis\IEEEauthorrefmark{2}, S{\l}awomir Sta\'{n}czak\IEEEauthorrefmark{1}\IEEEauthorrefmark{6}}

\IEEEauthorblockA{\IEEEauthorrefmark{1}Fraunhofer Heinrich Hertz Institute,  Germany, \{firstname.lastname\}@hhi.fraunhofer.de}
\IEEEauthorblockA{\IEEEauthorrefmark{4}Ericsson Research, Germany, \{alex.palaios, cara.watermann, philipp.geuer\}@ericsson.com}
\IEEEauthorblockA{\IEEEauthorrefmark{2}Vodafone Chair, Technische Universit\"{a}t Dresden, Germany, \{firstname.lastname\}@tu-dresden.de}
\IEEEauthorblockA{\IEEEauthorrefmark{3}Rheinland-Pf\"{a}lzische Technische Universit\"{a}t Kaiserslautern-Landau, Germany, \{partani, schotten\}@eit.uni-kl.de}
\IEEEauthorblockA{\IEEEauthorrefmark{5}Deutsche Telekom Chair, Technische Universit\"{a}t Dresden, Germany, \{firstname.lastname\}@tu-dresden.de}
\IEEEauthorblockA{\IEEEauthorrefmark{7}BMW Group, Germany, daniel.kuelzer@bmwgroup.com}
\IEEEauthorblockA{\IEEEauthorrefmark{6}Network Information Theory Group,
Technische Universit\"{a}t Berlin, Germany}
}

\maketitle

\begin{abstract}
The evolution of wireless communications into 6G and beyond is expected to rely on new \gls{ml}-based capabilities. These can enable proactive decisions and actions from wireless-network components to sustain \gls{qos} and user experience. Moreover, new use cases in the area of vehicular and industrial communications will emerge. Specifically in the area of vehicle communication, \gls{v2x} schemes will benefit strongly from such advances. With this in mind, we have conducted a detailed measurement campaign that paves the way to a plethora of diverse \gls{ml}-based studies. The resulting datasets offer
\acs{gps}-located wireless measurements across
diverse urban environments for both cellular (with two different operators) and sidelink radio access technologies, thus enabling a variety of different studies towards \gls{v2x}.
The datasets are labeled and sampled with a high time resolution.
Furthermore, we make the data publicly available with all the necessary information to support the on-boarding of new researchers. We provide an initial analysis of the data showing some of the challenges that \gls{ml} needs to overcome and the features that
\gls{ml} can leverage, as well as some hints
at potential research studies.  
\end{abstract}


\glsresetall

\section{Introduction}
\Gls{v2x} communication has a great potential to improve road safety, traffic efficiency, reduce congestions, and minimize energy consumption and emissions. It includes modalities such as \gls{v2v}, \gls{v2i}, \gls{v2n} and \gls{v2p} communication, and thus enables several \gls{cits} use cases such as collision warning, emergency vehicle warning, road works warning, cooperative lane change, platooning, and many others~\cite{v2xusecases}~\cite{kulzer2021ai4mobile}. These use cases have diverse network requirements in terms of latency, throughput, etc., and often require a certain level of \gls{qos} to function optimally. In order to fulfill these \gls{qos} requirements, it might be necessary to predict the network behavior. To that end, various \gls{ml} algorithms have been proposed~\cite{hao2018}.

Predicting \gls{qos}~\cite{v2xpqos} is a complex task since many different characteristics across different network layers can be targeted as \gls{qos}, e.g., received power in the physical layer or throughput in the application layer. There are multiple research works on these topics~\cite{moreira2020qos,mason2022,boban2021}. However, one common theme across the majority of these works is the lack of a curated dataset capturing both vehicle data and wireless network data that is readily available to the research community. Vehicle datasets focused on mobility traces using \gls{gps} and other sensor data are more common.
In~\cite{king2021survey}, a taxonomy to classify the open-source mobility traces based on data source, data collection methodology, and the mobility mode is presented. 

Datasets capturing both vehicle and network data require time, effort, and equipment to collect data samples in different vehicular environments such as cities, highways, rural roads, or a combination of them.  In~\cite{boban2022munich}, Boban et al. carried out a measurement campaign in Munich, Germany
to predict \gls{ul} throughput for \gls{v2x} scenarios such as \gls{tod} and local dynamic map update. The MOVESET dataset~\cite{chen2016moveset} presents a sensor package, based on correlated sensor readings with measurements of cellular and \gls{dsrc}, to help independent research groups create their own datasets. Another measurement campaign~\cite{qidataset} has been conducted on the interstate and rural highways in Bozeman, USA. The gatech/vehicular dataset~\cite{fujimoto2006gatech} presents collected data for short-range communication between vehicles, other vehicles, and roadside stations. In~\cite{cruz2017dsrc}, Cruz et al. carried out a measurement campaign in Porto, Portugal~\cite{cruz2020dataset} to improve the localization performance based on information from smartphone sensors and \gls{v2v} communication. In~\cite{torres2020lte}, Torres-Figueroa et al. verified the performance of the available \gls{lte} networks to deploy \gls{cits} applications safely. The measurement campaign was carried out in Munich, Germany for three different \glspl{mno} with a focus on \gls{qos} with respect to \gls{e2e} delays. In~\cite{mikami2020platooning}, Mikami et al. verified the dynamic mode switching for 5G NR-V2X sidelink communication for a vehicle platoon moving in and out of coverage on the Shin-Tomei expressway in Japan.

Previous datasets have been captured by conducting measurement campaigns in either an urban or a highway setting.
Our dataset~\cite{hernangomez2022berlinv2x} fills this gap in the literature with measurements from diverse vehicular environments mixing urban and highway scenarios. The dataset also includes network data from two different \glspl{mno}, which in turn can be used to assess whether the same \gls{ml} models can adapt from one \gls{lte} network to another. Moreover, our dataset provides high-resolution data for both cellular and sidelink communication, as well as metadata for weather and traffic conditions.

The remainder of the paper is organized as follows. \Cref{sec:measurement} explains the measurement campaign conducted in different vehicular environments in Berlin, Germany. \Cref{sec:data_analysis} provides a first analysis of the captured dataset. \Cref{sec:potential_studies} outlines the potential studies and, finally, a conclusion is given in \cref{sec:conclusion}.

\section{Measurement Campaign}\label{sec:measurement}
\begin{figure}[t]
    \centering
    \includegraphics[width=\linewidth]{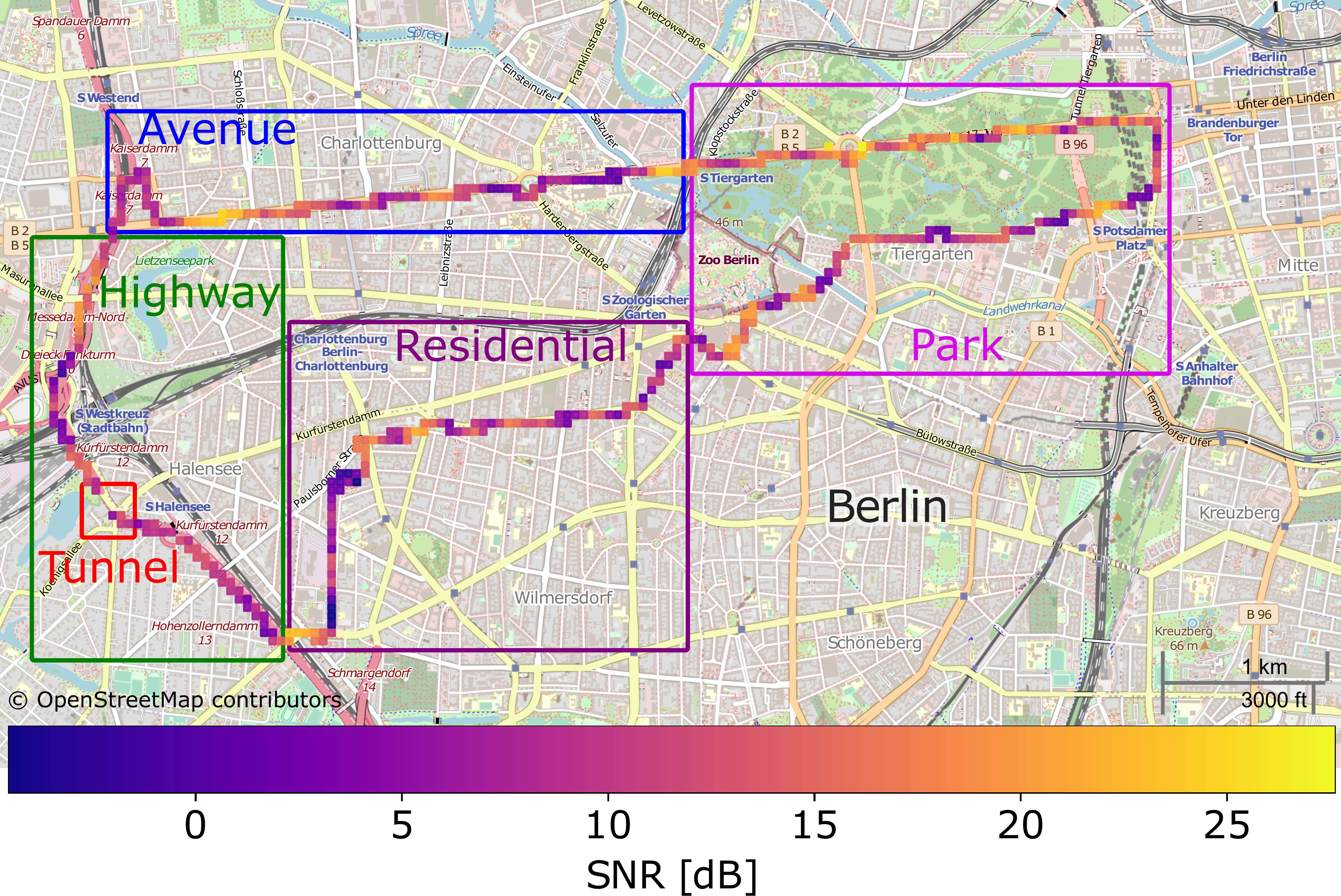}
    \caption{Measurement route in Berlin with \acs{snr} heatmap for operator 1.
    The different urban environments are highlighted.}
    \label{fig:map}
\end{figure}

Our data has been collected from drive tests around West Berlin along the route shown in \Cref{fig:map}.
This route contains a segment through the narrow \textit{residential} streets in the district of Wilmersdorf, broader \textit{avenues} in Charlottenburg and the central Tiergarten \textit{park}, and the westernmost segment of the urban \textit{highway} A100, between Hohenzollerndamm and Kaiserdamm, which also includes a short \textit{tunnel} beneath Rathenauplatz. The whole route stretches over \SI{17.2}{\kilo\meter} and a single drive round takes approximately \SI{45}{\minute} on a weekday morning.

Within three days, we have driven 17 clockwise rounds along the measurement route with up to four cars following one of two driving modes:

\begin{enumerate}
    \item Platoon: All vehicles in a convoy with fixed ordering (11 rounds).
    \item 2x2: Two pairs of vehicles driving approximately \SIrange{1.5}{3}{\kilo\meter} apart (6 rounds).
\end{enumerate}

We have measured two \glspl{rat} simultaneously:

\begin{enumerate}
    \item Cellular: A \gls{minipc} in each car exchanged data with a server over \gls{lte} and collected
    related measurements.
    The server was located in a data center in Berlin, and we used
    the public \gls{lte} network from two different \glspl{mno}
    on frequency bands between \SI{700}{\mega\hertz} and
    \SI{2.7}{\giga\hertz}. Two cars connected to 
    Vodafone's network and the other two to Deutsche Telekom. In the remainder of the paper,
    we alias them arbitrarily as operator 1 and 2 to shift the focus away from an explicit \gls{mno} comparison.
    \item Sidelink: Each car was also equipped with the \gls{sdr}
    platform described in \cite{lindstedt2020open}, which acted as a sidelink \gls{ue} according to \gls{3gpp} Release 14.
\end{enumerate}

For the cellular measurements, we reused the methodology from \cite{palaios2021network}, specifically the \gls{minipc} with the same modem, antenna, and \gls{gps} receiver. Also, the same applications including MobileInsight, Tcpdump and Iperf were used for data collection and generation. We considered low datarate scenarios at
\SI{400}{\kilo\bit\per\second} both for \glsfirst{ul} and \gls{dl} to obtain meaningful delay measurements, and performed experiments
at \SI{75}{\mega\bit\per\second} \gls{ul} 
and \SI{350}{\mega\bit\per\second} \gls{dl} 
to capture the maximum achievable throughput.
In each round, we assigned a specific
\gls{ul}/\gls{dl}  profile to each car and
labeled the transmission with a unique port to ease data preprocessing.

On the other hand, the sidelink data was collected according to the following operation regimes:
\begin{enumerate}[label=S\arabic*)]
    \item \Gls{cam} of length 69 bytes at a packet transmission rate of \SI{20}{\hertz} and \gls{mcs} 8 using two
    sub-channels (6 rounds).
    \item \Gls{cpm} of length 1000 bytes (including IP header and payload) at a rate of \SI{50}{\hertz} with \gls{mcs} 12 using ten sub-channels (5 rounds).
\end{enumerate}

In both cases, we used the frequency band of \SI{5.9}{\giga\hertz} and blind \gls{harq}.

Besides the wireless data, we also captured
the cars' GPS information and side
information including traffic conditions and
weather. \Cref{tab:data} provides
an overview of the captured data from
different sources with corresponding sampling intervals.

\begin{table*}[t]
\caption{Overview of the captured dataset}
\label{tab:data}
\begin{center}
\begin{tabular}{
p{0.1\textwidth}p{0.095\textwidth}p{0.08\textwidth}p{0.12\textwidth}p{0.52\textwidth}}
\hline
\textbf{Data category  }                   & \textbf{Source }                            & \textbf{Tool}                           & \textbf{Sampling interval} & \textbf{Features }                                                                                                    \\ \hline
\multirow{6}{*}{Cellular}         & \multirow{5}{*}{\glspl{minipc}}          & \multirow{3}{*}{MobileInsight} & \SI{10}{ms} & \Gls{phy}: SNR, RSRP, RSRQ, RSSI                                                                        \\
                                  &                                    &                                 & \SI{20}{ms}      &  Physical shared channels: assigned resource blocks, transport block size, \acs{mcs}                                                    \\
                                  &                                    &                                 & Event-based       & \Gls{rrc}: cell identity, \acs{dl}/\acs{ul} frequency, \acs{dl}/\acs{ul} bandwidth                                                         \\

                                  &                                    & Ping                            & \SI{1}{s}      & Delay                                                                                                \\
                                  &                                    & Iperf                           & \SI{1}{s}         & DL datarate, jitter                                                                                                  \\
                                  & Server                             & Iperf                           & \SI{1}{s}         & UL datarate, jitter                                                                                                  \\ \hline
Sidelink                          & \gls{sdr} platforms                      & TCPdump                         & Event-based       & SNR, RSRP, RSRQ, RSSI, noise power, \acs{rx} power, \acs{rx} gain                                            \\ \hline
Position                          & \gls{gps}                                &                                 & \SI{1}{s}         & Latitude, longitude, altitude, velocity, heading                                                             \\ \hline
\multirow{2}{*}{Side information} & \multirow{2}{*}{Internet database} & HERE API                        & \SI{5}{min}             & Traffic jam factor                                                    \\
                                  &                                    & DarkSky                         & \SI{1}{h}               & Cloud cover, humidity, precipitation intensity/probability, temperature, wind speed
\end{tabular}
\end{center}
\end{table*}

\section{Data analysis}\label{sec:data_analysis}
\begin{figure}[th]
\includegraphics[width=\linewidth]{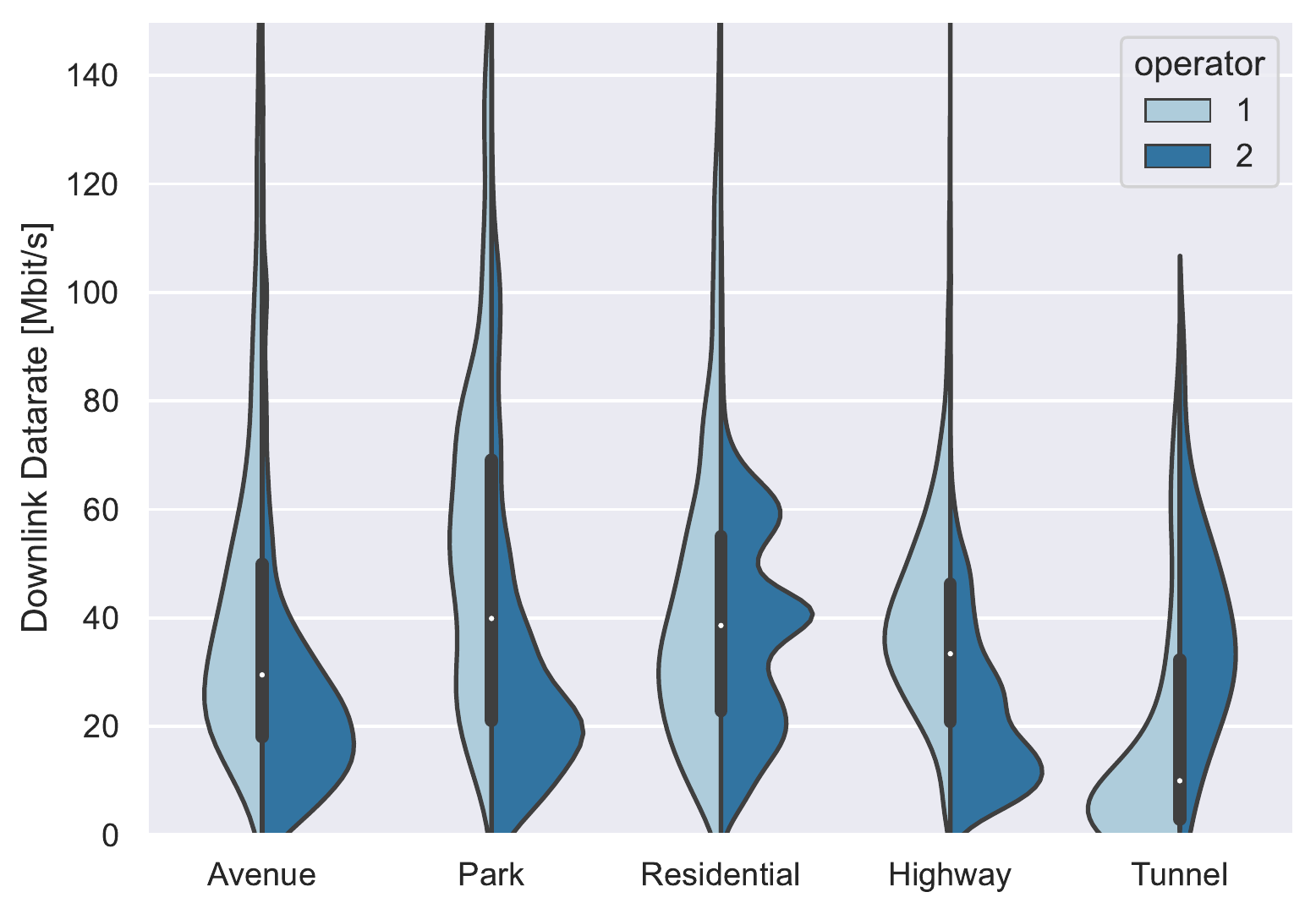}
\caption{Violin plots of downlink datarate for different areas.}
\label{fig:datarate_DL_violin}
\end{figure}

Here, we provide a preliminary analysis
of the dataset with a focus
on its diversity over geographical areas, 
operators, and devices.
The heatmap in \cref{fig:map} already hints at
some variation in the \gls{phy} characteristics along the route.
This effect is confirmed by
the datarate distribution for the different
measurement areas and operators in
\cref{fig:datarate_DL_violin}.
There, we can notice some similarities between operators,
e.g., for the avenue and the residential area.
This might be attributed to some extent to the characteristics of the specific radio environment and perhaps some reusing of the same sites from the operators,
which is a common strategy to reduce the cost of building and maintaining the infrastructure.
In other areas, such as the highway, the datarate distributions
seem to concentrate on disparate regions for each operator. Finally, the multimodal character of some distributions likely arises from the aggregation of two devices per operator.

\begin{figure}[t]
    \centering
    \includegraphics[width=\linewidth]{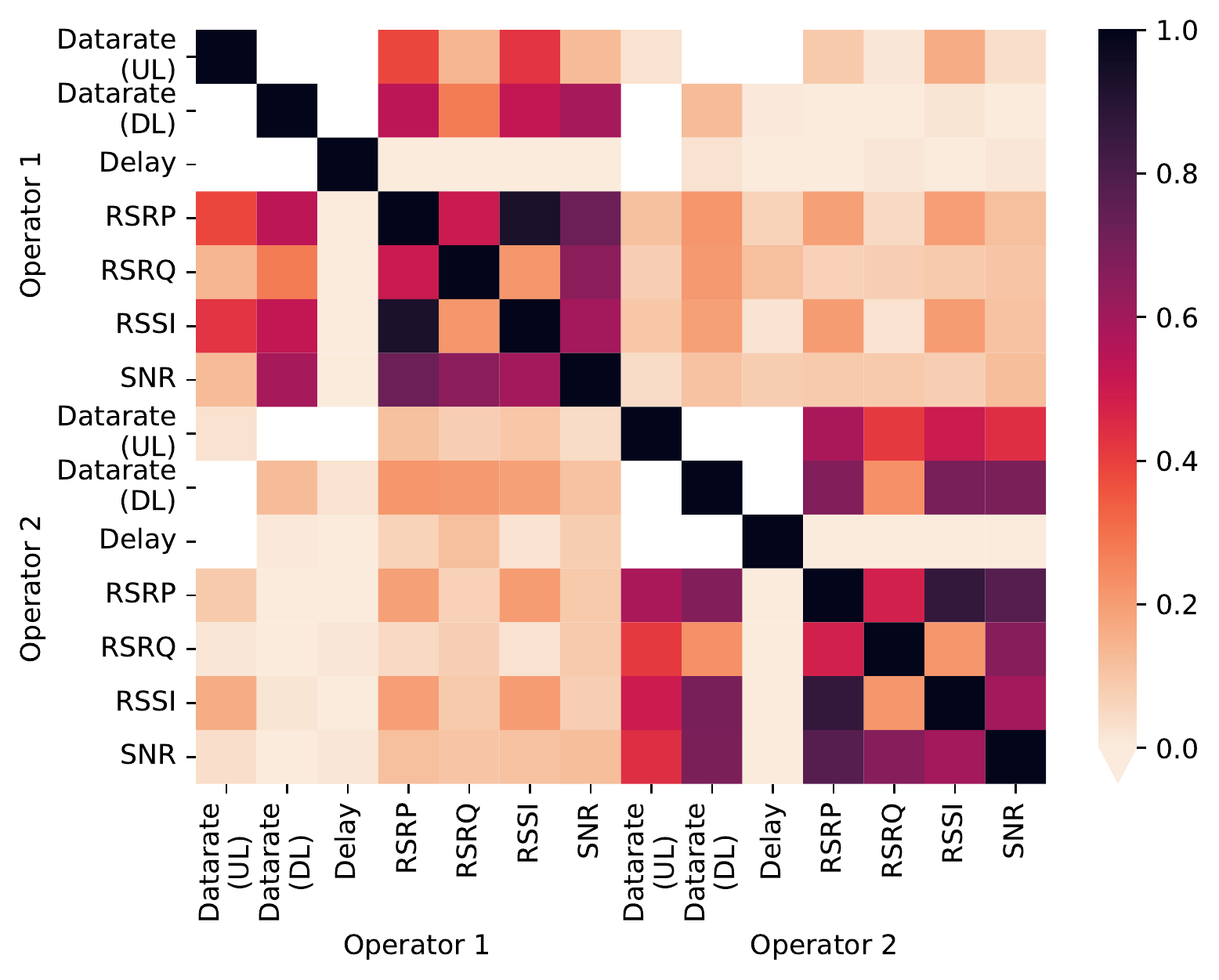}
    \caption{Correlation matrix for 2 devices from 2 operators.}
    \label{fig:corr_matrix}
\end{figure}

We continue with the study of the correlation between
features from both operators in \cref{fig:corr_matrix}. For each operator, we see that
the \gls{phy} features (most prominently the \gls{snr}) have a strong correlation with the
\gls{dl} datarate. For the \gls{ul} datarate, the
signal-strength type of
\gls{phy} features (\gls{rsrp} and \gls{rssi}) achieve
the largest correlation, especially for operator 1.
This discrepancy in \gls{ul} and \gls{dl} is expected,
since \gls{snr} is measured in the downlink.
The delay shows a weak correlation with
\gls{phy} features,
thus one should consider
additional information
for latency prediction.
A look into the cross-operator region of the matrix
reveals a modest correlation, most notably for \gls{rsrp} and \gls{rssi}. This, again, may be due to
site reuse across operators.

\begin{figure}[t]
  \centering
  \includegraphics[width=\linewidth]{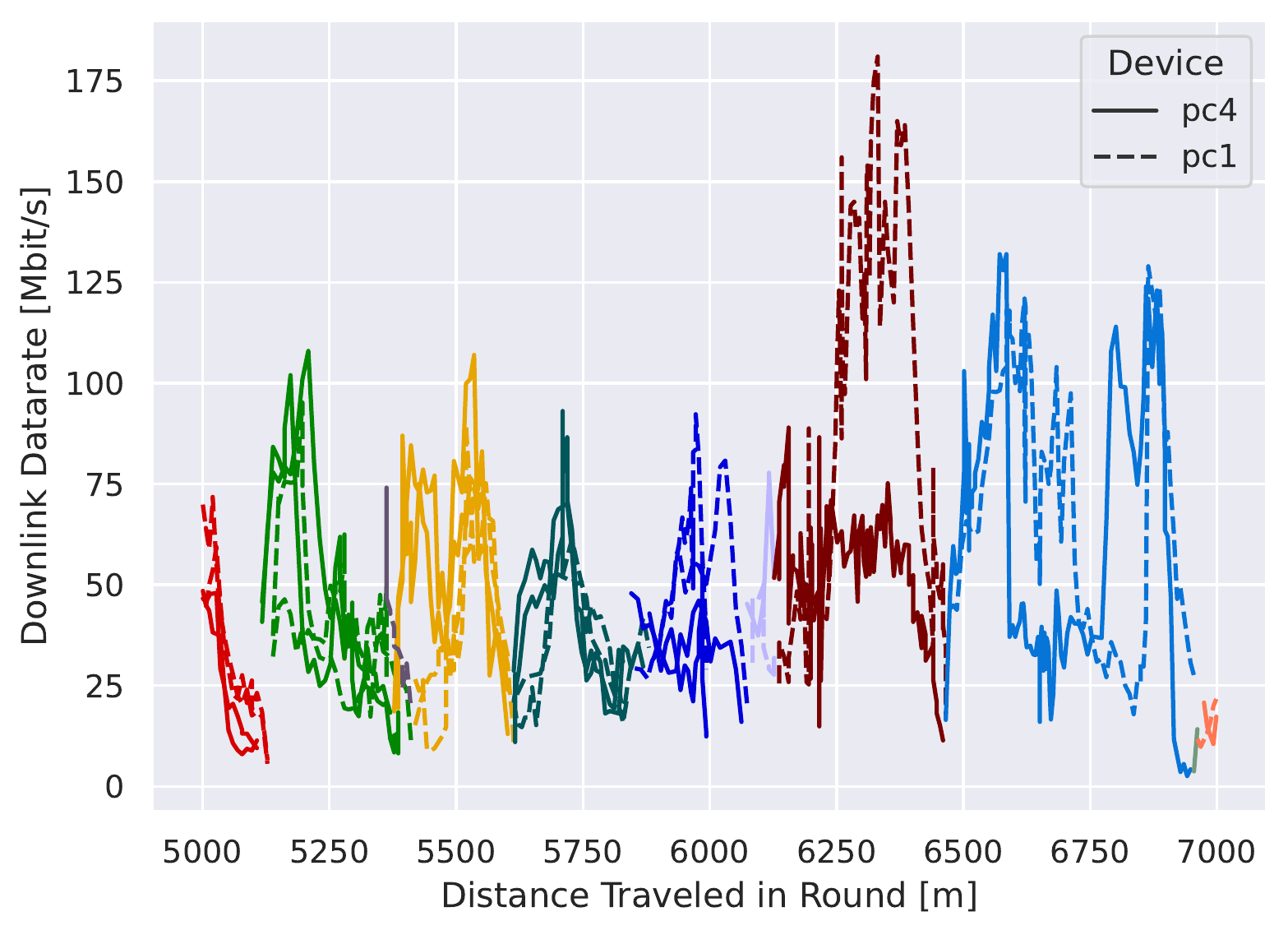}
\caption{Extract of downlink datarate along the round for operator 1. Color corresponds to cell identity.}
\label{fig:cell_plot_operator1}
\end{figure}

In \cref{fig:cell_plot_operator1}, we show the datarate and connected cell identity at different positions in the round for operator 1 along a
route segment of \SI{2}{\kilo\meter} within the
residential area. It can be clearly observed that both devices switch at roughly the same location to the same cells and the achievable datarates are lowest close to the handover points.
We have observed that this consistency in cell connectivity does not hold for operator 2 along the same route segment due to the fact that both devices use different frequency bands and thus have different cells available.

\begin{figure}[t]
    \centering
    \includegraphics[width=\linewidth]{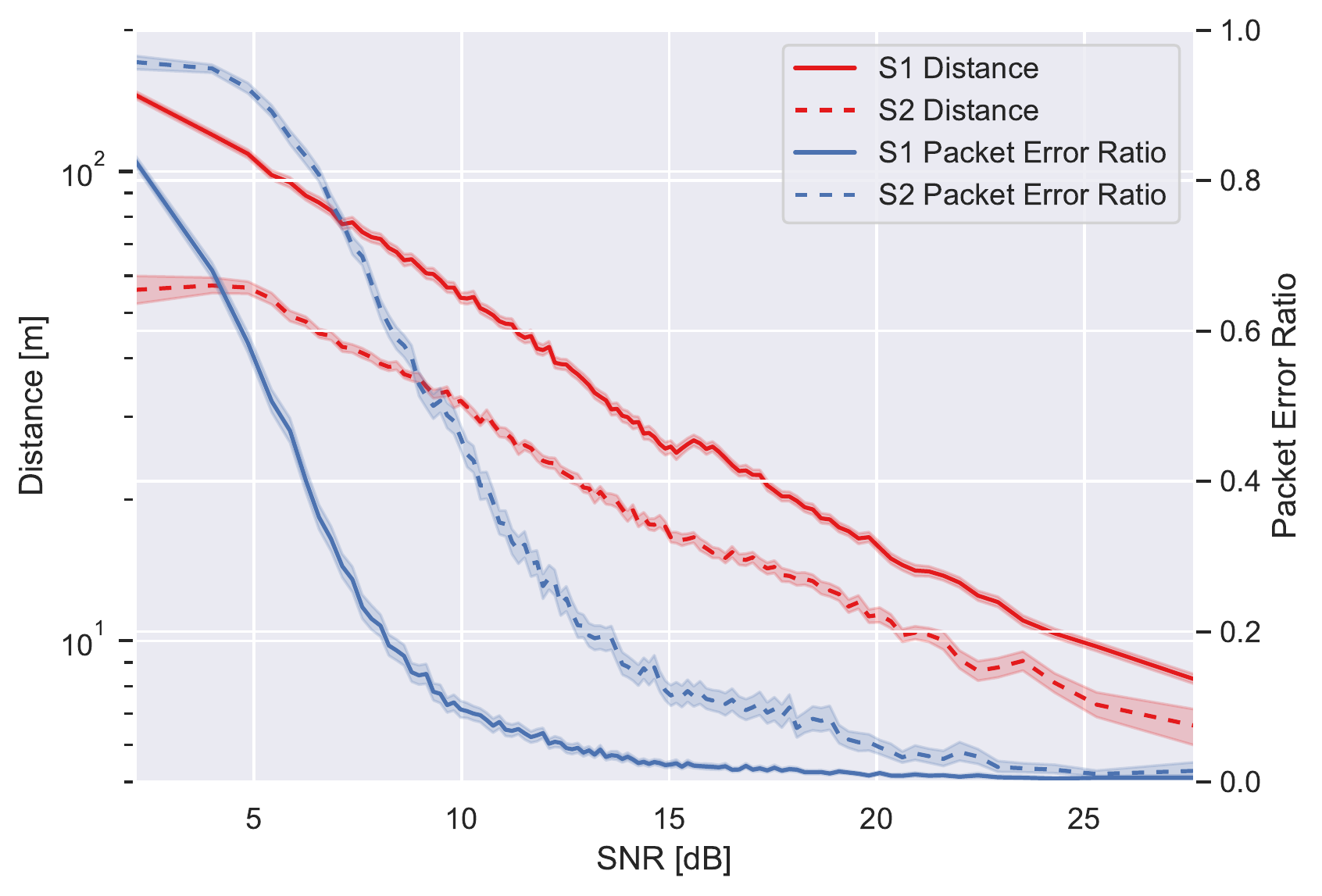}
    \caption{Relation of distance and sidelink \gls{per} with \gls{snr}.
    Several links have been combined for each line.}
    \label{fig:DistSnrPER}
\end{figure}

\begin{figure}[t]
    \centering
    \includegraphics[width=\linewidth]{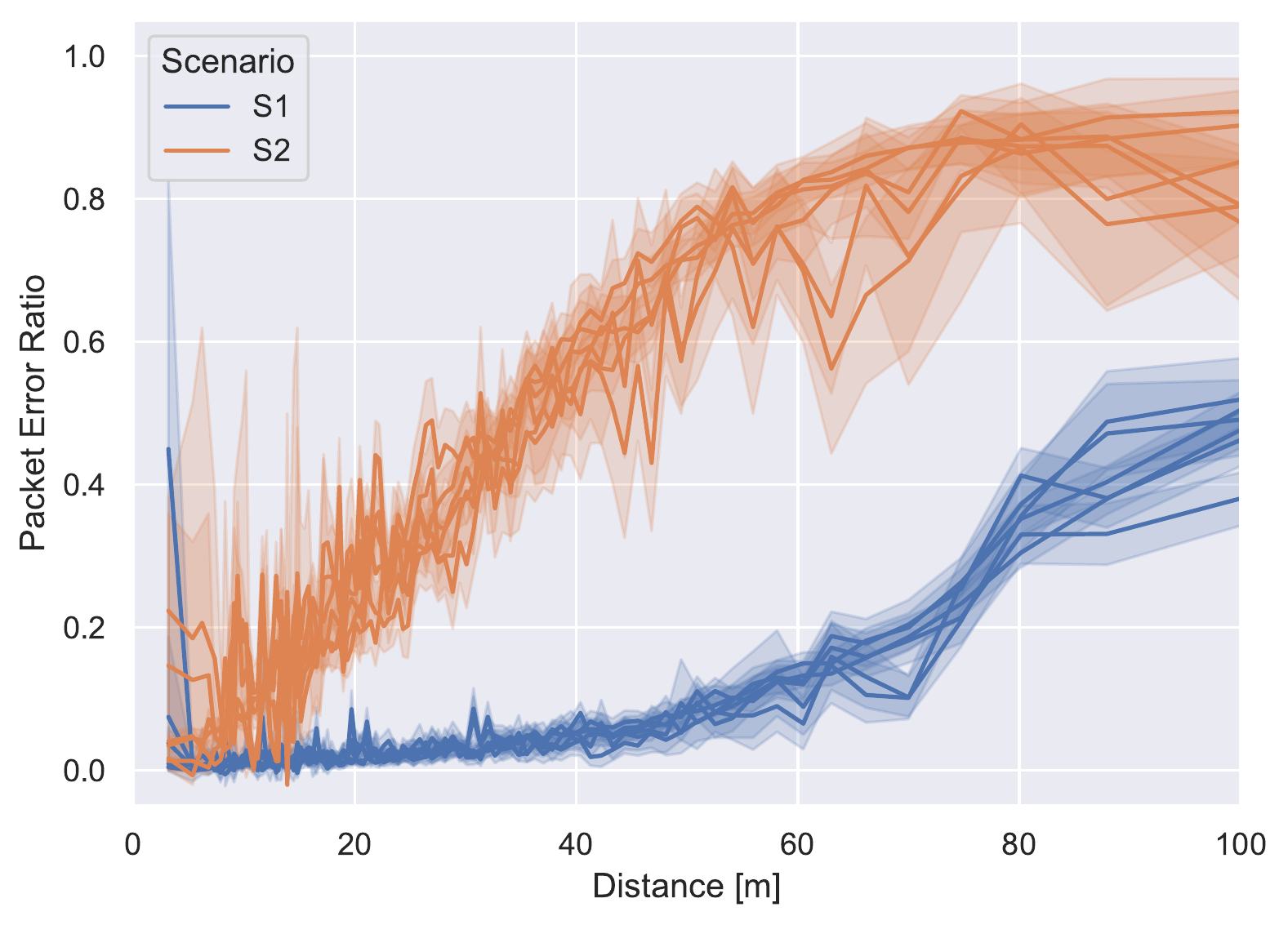}
    \caption{Relation of distance and \gls{per} with \SI{95}{\percent} confidence interval. Each line represents the sidelink between two
    \glspl{ue}.}
    \label{fig:DistPER}
\end{figure}

The \gls{gps} and sidelink measurements allow us to
investigate the practical link range for different
transmission modes as a relation between distance
and \gls{per}, which is mediated by the \gls{snr} (cf. \cref{fig:DistSnrPER}).
From a first
exploration of \cref{fig:DistPER}, one can estimate the link range as no more than \SI{80}{\meter} and
\SI{30}{\meter} for scenarios S1 and S2 defined
in \cref{sec:measurement}, respectively, and select transmission parameters
accordingly for the considered use case. Here, one should note that the employed
\gls{harq} mechanism provides extra tolerance to \gls{per}.

Due to the \gls{sdr}-based setup, the sidelink experiences some hardware-related effects.
Specifically, the transition of the power amplifier
from \gls{tx} to \gls{rx}
injects residual \gls{tx} power
in the adjacent \gls{rx} frames, which
leads to \gls{snr} degradation and a higher \gls{per}.
This holds true for all \glspl{ue},
but the so-called SyncRef \gls{ue} is most affected due to its periodic
transmission of synchronization sequences.
As a result, we have observed a different behavior
of the links to/from the SyncRef \gls{ue}
for the curves in
\cref{fig:DistSnrPER,fig:DistPER}, which
consequently
have been omitted from said figures.


\section{Potential Studies}\label{sec:potential_studies}
The high correlation between
datarate and \gls{phy} features, as shown
in \cref{fig:corr_matrix}, provides
a good starting point for
\gls{ml}-based \gls{qos} prediction, which
can be posed as various tasks thanks to the
measurement framework.
The captured cellular dataset also allows a comparison between two cellular networks deployed in the metropolis of Berlin from two different \glspl{mno}.
The understanding of the similarity between the two networks can contribute to previously conducted work in the area \cite{6817779}.
Moreover, the presented similarities already hint that \gls{ml} models can be reused, alleviating to some extent the need for recollecting data between \glspl{mno}.
The modest cross-correlation in
\cref{fig:corr_matrix} calls for
recent transfer learning techniques 
to bridge the gap between operators and devices.

The provided measurements may also be used to correct and calibrate simulations, which in turn can provide pre-training data for \gls{ml} models before their deployment
in a commercial network. Finally, such \gls{ml}-based prediction tasks can be thought of as an enabler for an increasingly
proactive \gls{rrm}, which leverages
predictions to react to upcoming changes in the wireless environment.

Moving on to sidelink communication, an interesting \gls{ml} research question is the estimation of \gls{per} from the distance
between devices.
\Cref{fig:DistPER} hints at a good estimation capability across a wide selection of links from \SIrange{5}{80}{\meter}.
Outside this range, \gls{gps} data does not provide enough accuracy for smaller distances and the data becomes unreliable for greater distances as \gls{per} approaches 1.
This relation clearly depends on settings such as packet rate or bandwidth, which is manifested by the differences between scenarios S1 and S2.

Finally, the inclusion of both cellular and sidelink data can help to test link selection strategies in a multi-\gls{rat} context. 

\section{Conclusion}\label{sec:conclusion}
In this paper, we present a complex dataset~\cite{hernangomez2022berlinv2x} containing \gls{gps}-located sidelink and cellular measurements from several devices, operators, and urban areas.
This enables not only multi-\gls{rat} \gls{qos} prediction but also transfer learning studies across different setups.

For the sidelink, the dataset provides insight into the achievable link range for two different settings.
Furthermore, we have observed that the relations between distance, \gls{snr}, and \gls{per} are consistent across several different links, which implies a good basis for \gls{ml}-based \gls{qos} predictors in the sidelink.

Testing \gls{ml} algorithms in such diverse scenarios can greatly increase confidence in \gls{ml} applicability for wireless vehicular communications.
In particular, model fine-tuning via online and transfer learning or domain adaptation can potentially reduce costly data collection procedures and thus improve \gls{ml} adoption.
The provided datasets allow extensive testing of the above-mentioned ideas.
Moreover, our preliminary analysis shows that the metropolitan area data is quite rich in dynamics and correlation patterns across operators and geographical areas, hence being a microcosmos to study many phenomena that are seen also in non-metropolitan setups. 

\bibliographystyle{IEEEtran}
\bibliography{IEEEabrv,bibliography.bib}

\end{document}